\documentclass[10pt,twocolumn,letterpaper]{article}

\usepackage[pagenumbers]{cvpr} 

\usepackage{lipsum}

\newcommand\blfootnote[1]{%
\begingroup
\renewcommand\thefootnote{}\footnote{#1}%
\addtocounter{footnote}{-1}%
\endgroup
}

\definecolor{cvprblue}{rgb}{0.21,0.49,0.74}
\usepackage[pagebackref,breaklinks,colorlinks,allcolors=cvprblue]{hyperref}
\usepackage{multirow}
\usepackage{amssymb}
\usepackage{fontawesome}
\usepackage{multirow}
\usepackage{pifont}
\usepackage{bbding}
\usepackage{caption}
\usepackage{lineno}

\title{Towards High-fidelity 3D Talking Avatar with Personalized Dynamic Texture}
\author{Xuanchen Li\textsuperscript{\rm 1}
$\quad$
Jianyu Wang\textsuperscript{\rm 1}
$\quad$
Yuhao Cheng\textsuperscript{\rm 1}
$\quad$
Yikun Zeng\textsuperscript{\rm 1}
$\quad$
Xingyu Ren\textsuperscript{\rm 1}\\
$\quad$
Wenhan Zhu\textsuperscript{\rm 3}
$\quad$
Weiming Zhao\textsuperscript{\rm 2}
$\quad$
Yichao Yan\textsuperscript{\rm 1\dag} \vspace{0.3em}\\
{\normalsize \textsuperscript{\rm 1}MoE Key Lab of Artificial Intelligence, AI Institute, Shanghai Jiao Tong University} \\
{\normalsize \textsuperscript{\rm 2}Student Innovation Center, Shanghai Jiao Tong University} $\quad$
{\normalsize \textsuperscript{\rm 3}Xueshen AI}\\
{\tt\small \{lixc6486, chengyuhao, yi1k-z, rxy\_sjtu, weiming.zhao, yanyichao\}@sjtu.edu.cn}\\
{\tt\small ppjiong11@alumni.sjtu.edu.cn, whzhu@foxmail.com}
}

\begin{document}
\twocolumn[{
\renewcommand\twocolumn[1][]{#1}
\maketitle
\begin{center}
    \captionsetup{type=figure}
    \includegraphics[width=\textwidth]{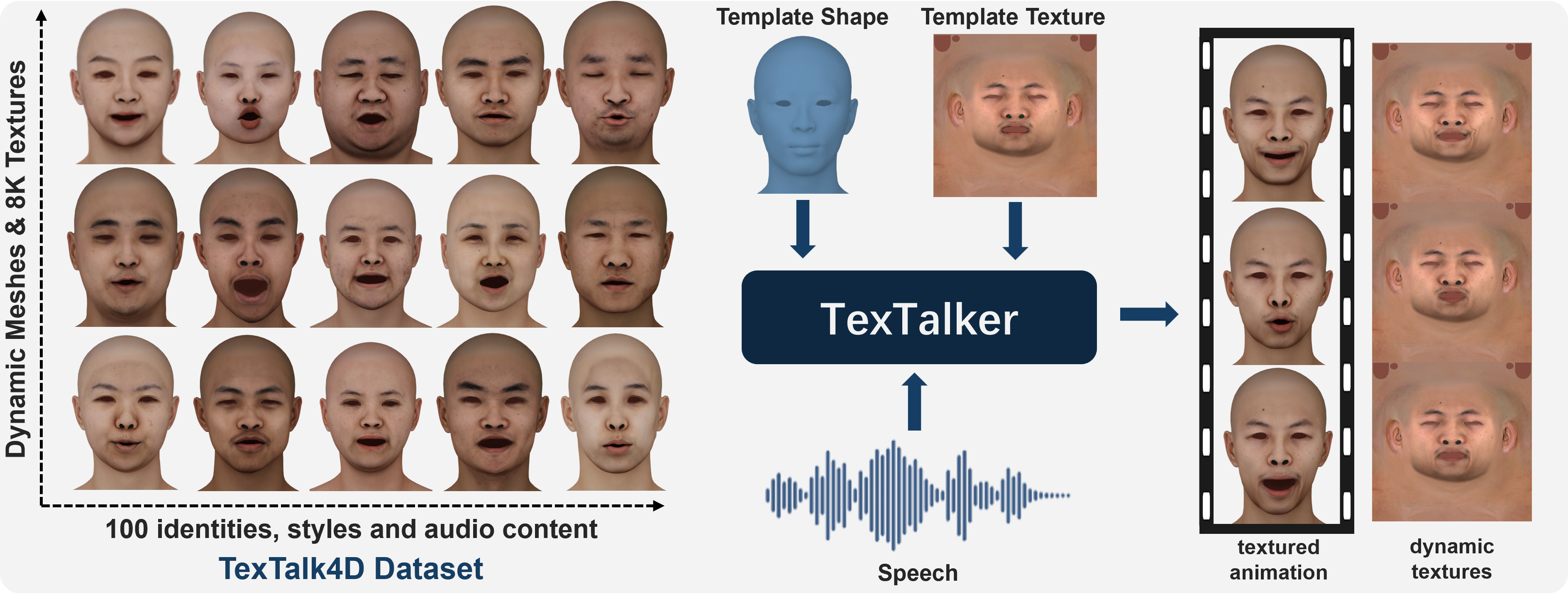}

    \captionof{figure}{We present \textbf{TexTalk4D}, a high-precision 4D audio-mesh-texture-aligned dataset consisting of 100 minutes of scan-level meshes with detailed 8K textures. Based on the dataset, we present \textbf{TexTalker} to generate geometry and aligned dynamic textures from speech simultaneously, advancing towards highly personalized textured facial animation.}

\end{center}
}]
\begin{abstract}
Significant progress has been made for speech-driven 3D face animation, but most works focus on learning the motion of mesh/geometry, ignoring the impact of dynamic texture. 
In this work, we reveal that dynamic texture plays a key role in rendering high-fidelity talking avatars, and introduce a high-resolution 4D dataset \textbf{TexTalk4D}, consisting of 100 minutes of audio-synced scan-level meshes with detailed 8K dynamic textures from 100 subjects. Based on the dataset, we explore the inherent correlation between motion and texture, and propose a diffusion-based framework \textbf{TexTalker} to simultaneously generate facial motions and dynamic textures from speech. Furthermore, we propose a novel pivot-based style injection strategy to capture the 
complicity of different texture and motion styles, which allows disentangled control. TexTalker, as the first method to generate audio-synced facial motion with dynamic texture, not only outperforms the prior arts in synthesising facial motions, but also produces realistic textures that are consistent with the underlying facial movements. Project page: \url{https://xuanchenli.github.io/TexTalk/}.
\end{abstract}    
\blfootnote{\textsuperscript{$\dag$}Corresponding author}
\section{Introduction}
Audio-driven 3D facial animation has been widely applied in entertainment media, such as movies and games, to generate vivid speech-aligned facial animation. The research in previous decades~\cite{voca, meshtalk, karras2017audio, faceformer, facediffuser, codetalker, diffposetalk} has focused on learning the correspondence between phonemes and facial motions while neglecting the impact of dynamic textures, 
greatly reducing the realism of the rendering results.
The industrial pipeline usually customizes compress and stretch maps manually created by professional artists to achieve dynamic textures~\cite{oat2007animated}, while it is too expensive and subject-specific to generalize to other models.
Therefore, there is an urgent demand to develop a general and robust method to effectively drive geometry together with dynamic texture.

To achieve this, one significant challenge
is the absence of high-quality dynamic texture datasets. 
The existing 4D datasets can be roughly divided into two categories based on their acquisition methods: those estimated from monocular videos~\cite{mmhead, media2face, peng2023emotalk} and those reconstructed from capture systems~\cite{voca, biwi, mmface4d, he2023speech4mesh, danvevcek2023emotional, meshtalk, zhang2019S3DFM, renderme360, multiface}.
Monocular estimated datasets can easily scale up by leveraging countless in-the-wild videos, but they fail to extract high-precision and temporal-consistent textures due to the uncertainties in uncontrolled environments. 
Meanwhile, the precision of capture systems and the high costs of data registration processes constrain high-quality dynamic texture acquisition.
Hence, few 3D talking head datasets contain dynamic textures. Although the Multiface dataset~\cite{multiface} contains dynamic textures, its quantity and quality are insufficient for general talking head model training.

Another challenge is that the co-generation of texture and geometry has not been well studied. Although previous works~\cite{facediffuser,diffposetalk,media2face} have achieved impressive results in speech-motion coarticulation, they cannot be trivially transferred to talking head generation with dynamic textures.
The primary obstacles lie in capturing the intricate long-term audio-visual correlations and maintaining the consistency between texture and geometry. Most previous works~\cite{voca, codetalker, faceformer, aneja2024facetalk} either directly generate vertex coordinates or parameterized coefficients~\cite{facediffuser, danvevcek2022emoca,danvevcek2023emotional}, which are extremely different from textures, resulting in difficulty in learning the complex relationship between texture and geometry.
Despite there being a recent work~\cite{media2face} using UV position maps to represent facial motion for better expressiveness, it does not explore motion-texture correlation. 

%

To tackle the aforementioned difficulties, we first propose a novel 3D talking head dataset with diverse dynamic textures, dubbed \textbf{TexTalk4D}, comprising 100 minutes and over 360,000 facial models. To construct this dataset, we
employ a high-precision capture system LightStage~\cite{debevec2012light} 
for head capture, and we utilize one of the SOTA dynamic face reconstruction methods Topo4D~\cite{li2024topo4d} to obtain high-quality head models and 8K high-resolution textures in the same topology automatically. 
Compared to existing datasets, TexTalk4D offers temporal-consistent dynamic textures with pore-level details, supporting the research on audio-driven textured 3D talking head generation.

Moreover, we propose \textbf{TexTalker}, a diffusion-based method for simultaneous driving of geometry and texture from audio. Specifically, \textbf{1)} we first represent the facial motions and texture variations as motion maps and wrinkle maps respectively to achieve a more unified representation of both. For the best of both expressiveness and efficiency, we then train two codebooks~\cite{vqgan, vqvae} to learn the facial animation primitives. \textbf{2)} Based on the learned low-dimensional spaces, we train a latent diffusion model~\cite{rombach2022high} to jointly capture the long-term motion-wrinkle correlation under speech guidance. \textbf{3)} Finally, to achieve disentangled motion-wrinkle style control, we propose a novel pivot-based disentangled style injection method. Benefiting from the expressive features stored in the animation primitive spaces, our method can effectively capture complex motion and wrinkling styles. Excessive experiments have proven that TexTalker not only achieves state-of-the-art geometry quality but also generates realistic and consistent dynamic textures. In summary, our contributions are as follows:

\begin{itemize}
  \item We propose a new task: audio-driven 3D talking head generation with corresponding dynamic textures.
  To fill the gap in the academic dataset for this task, we propose an open-source scan-level audio-geometry-texture-aligned dataset \textbf{TexTalk4D} captured from 100 subjects, with various styles and audio content.
  \item To the best of our knowledge,  we propose the first audio-driven geometry and texture method \textbf{TexTalker}, to jointly generate realistic facial motions together with consistently aligned dynamic textures.
  \item We propose a novel pivot-based style injection method that effectively captures the complex speaking and wrinkling styles, achieving disentangled style control.
\end{itemize}

\begin{figure*}[t]
  \centering
  \includegraphics[width=\linewidth]{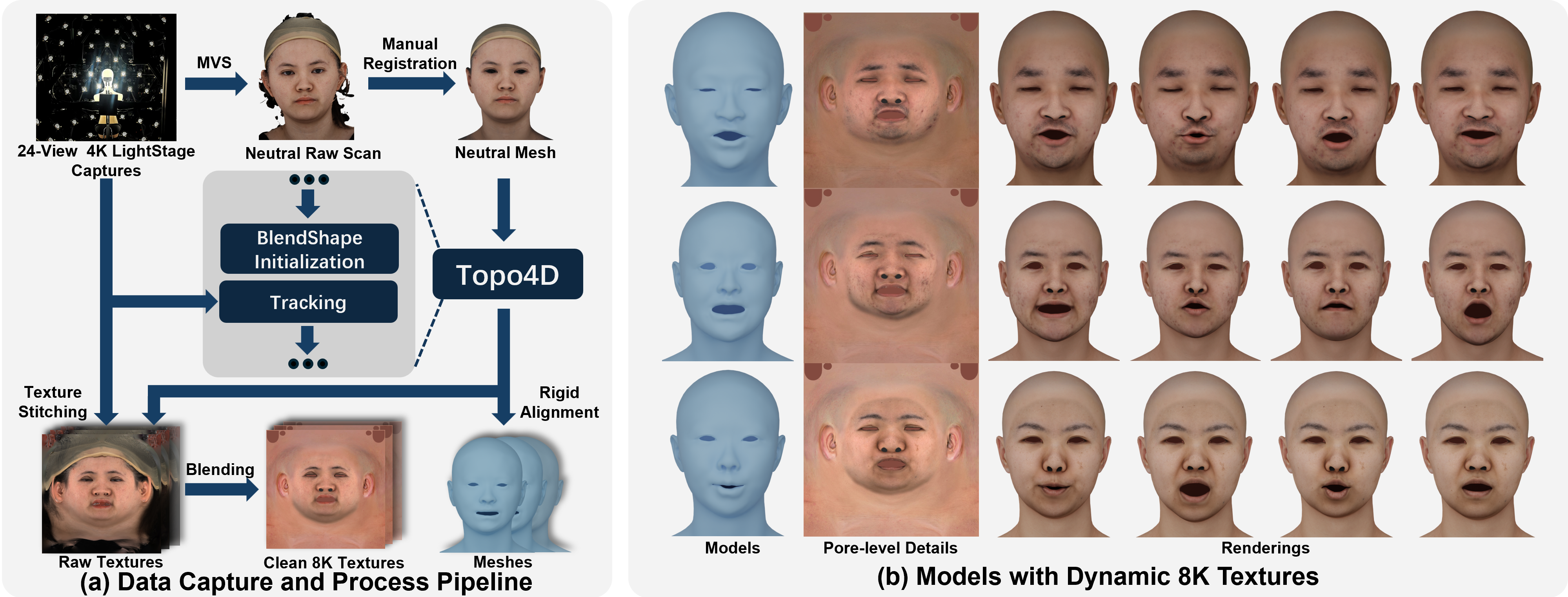}
  \caption{\textbf{Data processing pipeline.} We employ Topo4D~\cite{li2024topo4d} to obtain consistent meshes from LightStage captures. Based on the mesh sequence, we map the color from 24-view 4K images to get the 8K textures. Mesh alignment and texture blending are then conducted to get the final assets.}
  \vspace{-1em}
  \label{fig:dataset}
\end{figure*}
\section{Related Works}
\subsection{Speech-driven 3D Facial Animation}
Speech-driven 3D facial animation has been widely explored, and the major challenge in this task is to establish complex correlations between phonemes and vismes~\cite{fisher1968confusions}. 
Early works~\cite{kent1977coarticulation, edwards2016jali, cao2005expressive, deng2006expressive, wang2007assembling,taylor2012dynamic, xu2013practical} mostly segment speech into phonemes and build associations with vismes through rule-based or data-driven methods. However, these traditional methods always require a lot of manual operations from experts for rule design and parameter tuning, suffer high costs and fail to generalize to different languages. 
To achieve generalizable facial animation, learning-based methods attempt to exploit patterns between phonemes and audio features~\cite{baevski2020wav2vec, hannun2014deep, hubert}, which can be divided into two categories considering their animation representations. 
The first line of work employs implicit control signals, such as blendshapes and parameterized expression coefficients~\cite{blanz2023morphable, flame}, to drive face templates~\cite{busso2007rigid, pham2018end, pham2017speech, zhou2018visemenet, villanueva2022voice2face, facediffuser, diffposetalk, danvevcek2022emoca, peng2023emotalk, danvevcek2023emotional, ji2021audio}. However, these methods are limited by the quality of pre-defined expressions.
Another kind of method~\cite{voca, meshtalk, faceformer,codetalker,wu2023speech, thambiraja2023imitator, fan2022joint, chai2022personalized, chu2024corrtalk, yu2024ecavatar,facediffuser, aneja2024facetalk, karras2017audio, haque2023facexhubert,thambiraja2023imitator} gets rid of this limitation by
directly predicting vertex motions from audio, achieving more vivid results. Progressively, diffusion models~\cite{ho2020denoising, rombach2022high} are employed for better modeling the connection between audio and motions, involving various talking styles~\cite{facediffuser,diffposetalk, aneja2024facetalk,media2face, song2024expressive, chen2024diffsheg, mmhead}.
Despite generating expressive facial motions, these methods ignore dynamic facial texture, \eg, pore compression and wrinkle formation, while speaking, so that reduces realism and can even lead to an uncanny valley effect.

To obtain dynamic textures visually aligned with the facial motions, the traditional CG pipeline typically constructs a custom dynamic texture library~\cite{oat2007animated} from the identity-specific expression bases, so that the facial texture changes can be sampled from the library according to the vertex offsets. However, this method requires expensive assets and has a huge computational cost, hard to generalize to different identities. 
Based on generated facial motions, some methods~\cite{animateme, li2020dynamic, p2phd} can generate Action-Unit-aware~\cite{ekman1978facial} dynamic textures, but they ignore temporal consistency. Besides, because geometry and texture are inherently strongly correlated, fully utilizing their complementary information should promote mutual generation. However, few previous research have explored this issue. To unify the generation process of geometry and texture, TexTalker learns compact latent spaces to represent facial motions and texture changes, enabling audio-driven facial motion and corresponding dynamic texture generation.

\subsection{Existing 4D Audio-Visual Dataset}
\begin{table}[t]
\centering
\resizebox{\columnwidth}{!}{
\begin{tabular}{lccccc}
\toprule
\multirow{2}{*}{Dataset}           & Capture & Identity & Temporal & Pore-level & Texture\\ 
 & Env. & Number & Consistency & Details &Resolution\\
\toprule
3D-ETF~\cite{peng2023emotalk}               & \multirow{3}{*}{Mix} & 100 & \XSolidBrush                & \XSolidBrush       & -                  \\
MMHead~\cite{mmhead}               &  & 2000 & \XSolidBrush                & \XSolidBrush       & -                  \\
M2F-D~\cite{media2face}                &                      & 500  & \XSolidBrush                & \XSolidBrush       & -                  \\ \hline
BIWI~\cite{biwi}                 & \multirow{9}{*}{Lab} & 20   & \XSolidBrush                & \XSolidBrush       & -                  \\
S3DFM~\cite{zhang2019S3DFM}              &                      & 77   & \XSolidBrush                & \XSolidBrush       & -                  \\
VOCASET~\cite{voca}              &                      & 12   & \XSolidBrush                & \XSolidBrush       & -                  \\
MeshTalk~\cite{meshtalk}              &                      & 250   & \XSolidBrush                & \XSolidBrush       & -                  \\
MEAD-3D~\cite{danvevcek2023emotional, he2023speech4mesh}              &                      & 60   & \XSolidBrush                & \XSolidBrush       & -                  \\
MMFace4D~\cite{mmface4d}             &                      & 431  & \XSolidBrush                & \XSolidBrush       & -                  \\
RenderMe-360~\cite{renderme360}         &                      & 500  & \XSolidBrush                & \XSolidBrush       & $256$     \\
Multiface~\cite{multiface}            &                      & 13   & \Checkmark                  & \Large\textcolor{black}{\ding{51}}{\large\textcolor{black}{\kern-0.77em\ding{55}}}                   & $1024$   \\
\textbf{TexTalk4D (Ours)} &                      & 100  & \Checkmark                  & \Checkmark         & $8192$   \\ \bottomrule
\end{tabular}
}
\captionsetup{singlelinecheck=false}
\caption{\textbf{Existing Audio-visual datasets comparison.} TexTalk4D contains \textbf{temporal-consistent 8K textures} with realistic \textbf{pore-level details}, surpassing existing datasets.}
\label{tab:datasets}
\end{table}
There have been many audio-visual datasets available, as is shown in Tab.~\ref{tab:datasets}, including those\cite{mmhead, media2face, peng2023emotalk} extracted from in-the-wild videos and those~\cite{voca, biwi, mmface4d, he2023speech4mesh, danvevcek2023emotional, meshtalk, zhang2019S3DFM, renderme360, multiface} preciously reconstructed from captures. 
Although datasets reconstructed from videos can achieve astonishing diversity in identity and speech, they cannot faithfully reconstruct subtle facial expressions or extract temporal consistent, and detailed textures. 
On the other hand, the cost of producing 3D datasets is enormous because of the limitations of capture equipment, labor, and computational costs, they are limited in scale and diversity, making it struggle to meet the training needs. 
Currently, only RenderMe-360\cite{renderme360} and Multiface\cite{multiface} provide regular-UV textures. However, the resolution of RenderMe-360 is low, with obvious noise and poor temporal consistency. The one closest to ours is Multiface, but it lacks diversity and the details in the textures are limited. To fill the gap in the academic community's lack of 4D datasets with high-quality textures, we propose the \textbf{TexTalk4D}, which consists of rich speech-aligned scan-level assets from 100 subjects, especially clean 8K dynamic textures with pore-level details.

\section{TexTalk4D Dataset}

\subsection{Data Acquisition}
As is shown in Fig.~\ref{fig:dataset}, We use a LightStage~\cite{lightstage} with 24 synced cameras to capture high-precision talking data. Please refer to Sup.~Mat. for more details about the capture system. For the sake of efficiency and accuracy, we use Topo4D~\cite{li2024topo4d}, one of the most advanced 4D reconstruction pipelines, to acquire the temporal-consistent facial models from multi-view videos. 
The first frame of facial geometry and texture are manually registered~\cite{goesele2006multi, besl1992method} for the best tracking quality. 
Since the vanilla Topo4D struggles to track high-speed eye blinking, we improved it with facial priors. 
Specifically, we create the eye-blink blendshape for the models in our topology, which is general across different identities.
Before tracking each frame, we estimate the weight of the eye-blink with Mediapipe~\cite{lugaresi2019mediapipe} and initialize the vertices of the eyes using the pre-defined blendshape. Although the estimation is not accurate, the optimization process of Topo4D will start from the approximately correct initialization. Therefore, benefiting from the excellent performance of Topo4D and the relatively precise initialization from facial priors, our reconstructed meshes achieve high accuracy.
After acquiring meshes, we map the 24-view 4K images to UV space and stitch them into texture maps.
For better visualization and reducing artifacts, we remove the redundant areas in the raw textures by alpha blending with template texture maps. To achieve better texture quality, we apply the skin tone transfer method to the template for seamless blending.
Finally, all the reconstructed meshes are rigidly aligned with the template mesh automatically.
\subsection{Data Statistics}
The proposed dataset consists of audio-mesh-texture pairs from 100 Asian youth, including 60 males and 40 females. Both the meshes and textures are topology-consistent and temporal-consistent.
The meshes contain the whole head, composed of 8280 vertices and 16494 faces, while the textures are 8K resolution. 
For each subject, we collect one minute of Mandarin speech data at 60FPS. To maximize the phonetic diversity, we do not strictly require the speech content, but allow the subjects to freely prepare the text.
\section{Methodology}
\begin{figure*}[t]
  \centering
  \includegraphics[width=\textwidth]{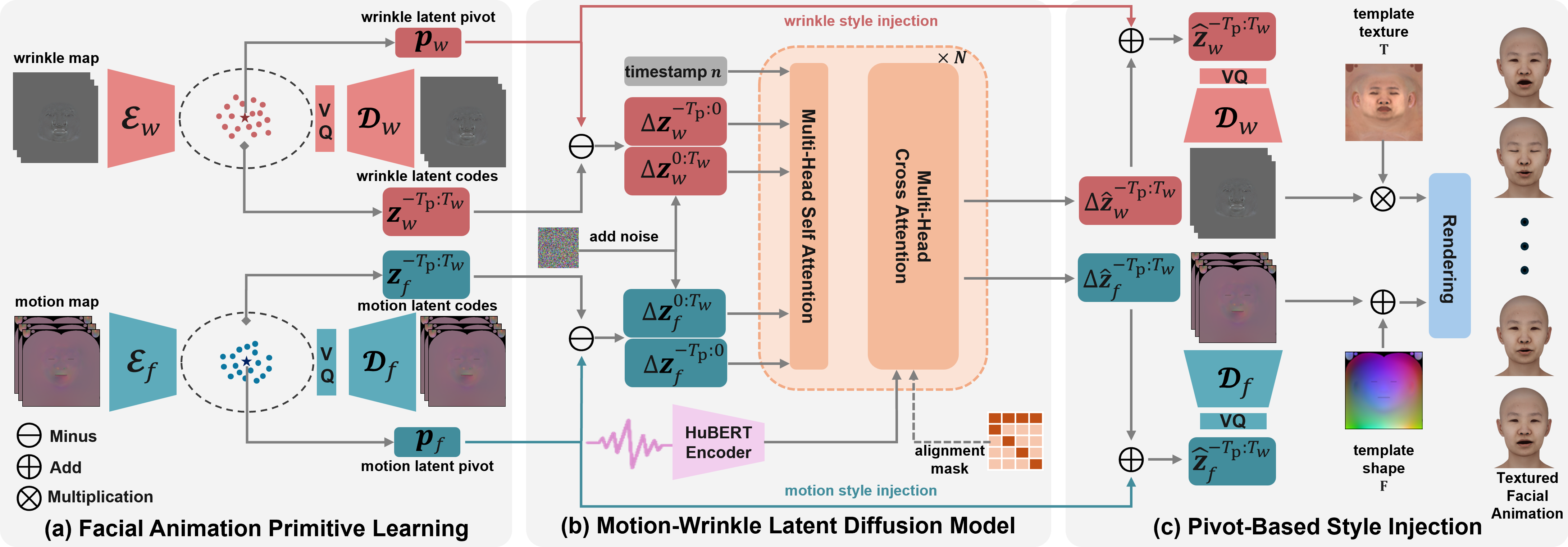}
\caption{\textbf{The overview of TexTalker.} \textbf{(a)} We train quantized autoencoders to unify the representation of geometry and texture with better efficiency. (b) Based on the learned low-dimensional animation primitives, we employ an LDM to jointly diffuse geometry and texture latent offsets $\Delta \mathbf{z}$ from the style pivots $\mathbf{p}$ for long-term correlation learning. (c) By adding back the style pivots, the motion and wrinkle styles can be independently controlled. Finally, the personalized textured animation assets can be obtained by decoders.}
\vspace{-1em}
  \label{fig:pipeline}
\end{figure*}
\subsection{Task Formulation}
Our goal is to generate personalized textured facial animation from arbitrary speech. Specifically, given $\mathbf{T}$ frame speech $\mathbf{}{A}_{1:T}=(\mathbf{a}_1,...,\mathbf{a}_T)$, where each $\mathbf{a}_{t} \in \mathbb{R}^d$ has $d$ samples, our model can generate facial motions $\mathbf{M}_{1:T}=(\mathbf{m}_1,...,\mathbf{m}_T)$ and texture variation $\mathbf{W}_{1:T}=(\mathbf{w}_1,...,\mathbf{w}_T)$, 
where $\mathbf{m}_t \in \mathbb{R}^{n_f\times 3}$ is the vertex offset from the cannonic facial model and $\mathbf{w}_t \in \mathbb{R}^{H\times W\times3}$ means the per-pixel ratio from the texture in neutral expression. In addition, we expect to control the styles of facial motions $\mathbf{M}_{1:T}$ and texture variation $\mathbf{W}_{1:T}$ separately.

To this end, we propose \textbf{TexTalker}, as illustrated in Fig.~\ref{fig:pipeline}. Firstly, we represent facial motions and texture variation as motion maps and wrinkle maps, and then respectively compress them into two independent low-dimensional latent spaces (Sec. \ref{sec: vqgan}). Based on the compact facial animation representation, we design a latent diffusion~(LDM)-based model to synthesize the animation primitives stored in the codebooks (Sec. \ref{sec: LDM}). Finally, we novelly extract style pivot $\mathbf{p}_m$ and $\mathbf{p}_w$ representing various styles from trained latent spaces, and then inject them into the outputs of LDM to achieve decoupled speaking and wrinkling style control (Sec. \ref{sec: stylize}).

\subsection{Facial Animation Primitive Learning}
\label{sec: vqgan}
As discussed in the introduction, it is essential to represent facial motion and texture variation in similar spaces for better learning cross-domain alignment and correlation. Therefore, we first transfer 3D models to UV space as textures.
Specifically, we map vertex offsets $\mathbf{m}_{t}$ to UV space according to the UV coordinates as the motion map~$\mathbf{f}_{t}=\mathcal{R}(\mathbf{m}_t)$, where $\mathcal{R}$ stands for the UV mapping operation.
Since identity information and expressions are coupled in the texture maps, we follow~\cite{zhang2022video} to represent texture variations by wrinkle map $\mathbf{w}_t$ to get rid of bias between speech and identities in training data.

Despite the collaborative representation of geometry and texture, directly training the generative model on UV maps leads to considerable computational costs and difficulty.
The recent success of discrete prior representation~\cite{vqvae, mentzer2023finite} inspires us to compress the UV maps to a low-dimensional latent space. The expressive latent space provides realistic facial animation primitives and reduces cross-modal ambiguity~\cite{codetalker}, and the following animation generation network only needs to learn animation primitives at a low cost. 
Hence, we train two independent quantized auto-encoders $\mathcal{E}_f$ and $\mathcal{E}_w$ in the same way to compress discrete motion and wrinkle primitives, respectively. Following VQGAN~\cite{vqgan}, taking motion for example, we jointly train the motion encoder $\mathcal{E}_f$, codebook $\mathcal{C}_f$, generator $\mathcal{G}_f$ and discriminator $\mathcal{D}_f$, where $\mathcal{E}_f$ compresses the map $\mathbf{f}$ into a low-dimensional latent $\mathbf{z}_f=\mathcal{E}_f(\mathbf{f}) \in \mathbb{R}^{h\times w \times d}$. Then each grid vector in the continuous $\mathbf{z}_f$ is translated by finding the nearest vector in the codebook $\mathcal{C}_f=\{\mathbf{c}_{f}^n \in \mathbb{R}^d\}_{n=1}^C$ through a quantization operation $\mathcal{Q}_f(\cdot)$:
\begin{equation}
      \tilde{\mathbf{z}}_f^{(i,j)} = \mathcal{Q}_f(\mathbf{z}_f^{(i,j)}) := \arg\mathop{min}_{c_f^n\in\mathcal{C}_f}\Vert \mathbf{z}_f^{(i,j)} - c_f^n \Vert.
\end{equation}

Finally, the generator $\mathcal{G}_f$ can reconstruct the map $\hat{\mathbf{f}}=\mathcal{G}_f(\tilde{\mathbf{z}}_f)$.
The training objective consists of (1) $L_1$ reconstruction loss $\mathcal{L}_{\text{rec}}$, (2) VGG~\cite{vgg} perceptual loss $\mathcal{L}_{\text{per}}$, (3) adversarial gan loss $\mathcal{L}_{\text{adv}}$ and (4) codebook loss $\mathcal{L}_{\text{code}}$. 
To sum up, the training objective can be formulated as:
\begin{equation}
    \mathcal{L}_{\text{latent}} = \mathcal{L}_{\text{rec}} + \eta_{\text{per}}\mathcal{L}_{\text{per}} + \eta_{\text{adv}}\mathcal{L}_{\text{adv}} + \eta_{\text{code}}\mathcal{L}_{\text{code}}.
\end{equation}



\subsection{Motion-Wrinkle Latent Diffusion Model}
\label{sec: LDM}
After unifying the representation,  we train a transformer-based latent diffusion model~(LDM) $\mathcal{F}$ to learn the correlation between facial motion latent $\mathbf{z}_f$ and wrinkle latent $\mathbf{z}_w$ under the guidance of audio $\mathbf{A}$. 
Specifically, $\mathbf{z}_f$ and $\mathbf{z}_w$ are firstly flattened and concated together to form a motion-wrinkle sample $\mathbf{X}^0 = [\mathbf{z}_f, \mathbf{z}_w]$. The clean sample $\mathbf{X}^0$ is then added noise $\boldsymbol{\epsilon} \sim \mathcal{N}(\mathbf{0}, \mathbf{I})$ as $\mathbf{X}^n (n=1, 2, ..., N)$ for $n$ times iteratively. The LDM is to denoise the $\mathbf{X}^n$ gradually guided by the audio features extracted by a pre-trained HuBERT~\cite{hubert} audio encoder. Notably, an alignment mask~\cite{faceformer, diffposetalk} is used for audio-motion-wrinkle alignment, so that the motion-wrinkle features are only associated with the audio features of the same frame. As is shown in Fig.~\ref{fig:pipeline}, for the learning of long-term dependence and more temporal consistent results, the diffusion process is conducted in a window range~\cite{diffposetalk} as:
\begin{equation}
\label{equ: diffusion}
    \hat{\mathbf{X}}^0_{-T_p:T_w} = \mathcal{F}(\mathbf{X}^n_{0:T_w},\mathbf{X}^0_{-T_p:0},\mathbf{A}_{-T_p:T_w},\mathbf{n}),
\end{equation} where the network simultaneously synthesise the clean samples $\hat{\mathbf{X}}^0_{-T_p:T_w}$ of current window of length $T_w$ and previous window of length $T_p$. The clean samples $\mathbf{X}^0_{-T_p:0}$ from the previous window and a diffusion timestep $\mathbf{n}$ are also fed in for better guidance.


The simple loss~\cite{ho2020denoising} is utilized for training the motion-wrinkle LDM as:
\begin{equation}
    \mathcal{L}_\mathcal{F} = \Vert \hat{\mathbf{X}}^0_{-T_p:T_w} - \mathbf{X}^0_{-T_p:T_w}\Vert^2.
\end{equation}
Based on the LDM, we can generate audio-synced facial motions $\hat{\mathbf{f}}=\mathcal{G}_f(\mathcal{Q}_f(\hat{\mathbf{z}}_f))$ with dynamic wrinkles $\hat{\mathbf{w}}=\mathcal{G}_w(\mathcal{Q}_w(\hat{\mathbf{z}}_w))$, thus the sequence of 3D facial meshes and dynamic textures can be recovered.

\subsection{Pivot-Based Disentangled Style Injection}\begin{figure}[t]
  \centering
  \includegraphics[width=\linewidth]{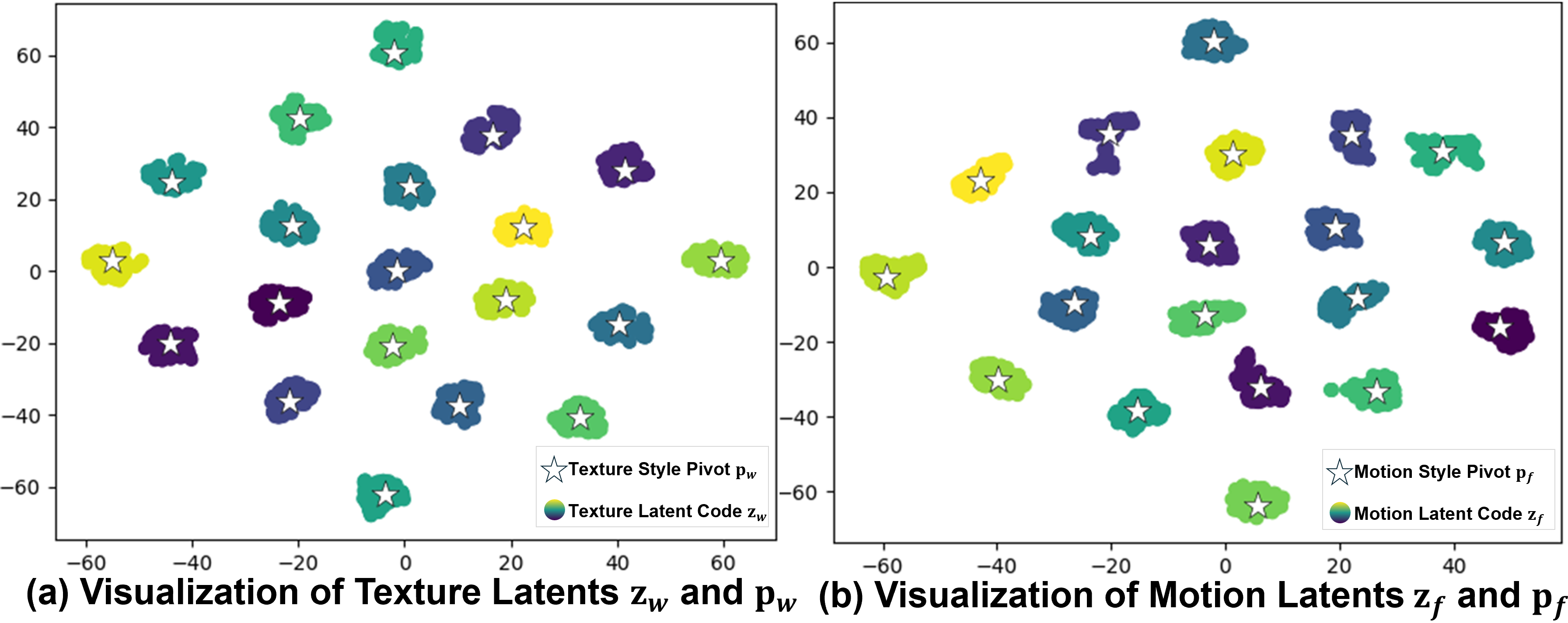}
  \caption{\textbf{t-SNE distribution of latent features from 20 subjects.} The learned animation primitive spaces effectively distinguish latent features of different styles both in wrinkles and motions. }
  \vspace{-1em}
  \label{fig:latents}
\end{figure}
\label{sec: stylize}
To further achieve style control, previous works mostly adopt one-hot style embedding~\cite{voca,codetalker,faceformer,facediffuser} or style feature extracted from examples~\cite{diffposetalk, thambiraja2023imitator, yang2024probabilistic}. Despite its simplicity, one-hot embedding fails to capture complex styles and is short of generalization ability. Besides, previous example-based methods usually require fine-tuning or training additional feature extraction networks. For the best of both simplicity and expressiveness, we propose a novel style injection method, where the style can be directly extracted from the learned animation primitives. 

Our key insight is that the learned codebook stores the animation primitives, and facial animation with similar style shares similar primitives, thus the latent corresponding to the same style should be clustered together. As shown in Fig.~\ref{fig:latents}, we observe that the motion latent $\mathbf{z}_f$ and wrinkle latent $\mathbf{z}_w$ from the same subject (whether it is in the training set or unseen) are clustered and the centroid of the cluster, \ie, style pivot $\mathbf{p}$, covers the main features of the animation style, which can be calculated by the average of a sequence of latent codes $\mathbf{z}_{1:T}$ from the same identity:
\begin{equation}
    \mathbf{p} = \frac{1}{T}\sum_{t=1}^T \mathbf{z}_t.
\end{equation}

Notably, a unique advantage of the style pivot is that it comes from the same space as the facial animation latent codes. Our animation generation model thus only needs to predict phoneme-relevant but style-free latent offsets from the style pivot and the style can be seamlessly injected, achieving disentangled style control. Specifically, since the animation latent $\mathbf{z}$ and style pivot $\mathbf{p}$ are both style relevant, we hypothesise that the offset from $\mathbf{p}$ to $\mathbf{z}$, \textit{i.e.}, $\Delta \mathbf{z} = \mathbf{z} - \mathbf{p}$, is style free but only phoneme relevant. Formally, the motion-wrinkle sample is defined as $\mathbf{X}^\prime_0 = [\Delta \mathbf{z}_f, \Delta \mathbf{z}_w]$, and the diffusion process (Eq.~\ref{equ: diffusion}) can be reformulated as:
\begin{equation}
    \hat{\mathbf{X}}^{\prime 0}_{-T_p:T_w} = \mathcal{F}(\mathbf{X}^{\prime n}_{0:T_w},\mathbf{X}^{\prime 0}_{-T_p:0},\mathbf{A}_{-T_p:T_w}, \mathbf{n}),
\end{equation}
where the LDM $\mathcal{F}$ predicts the latent offsets $\Delta \hat{\mathbf{z}}_f$ and $\Delta \hat{\mathbf{z}}_w$. Further, the wrinkle and motion latent can be obtained by adding back a style pivot $\mathbf{p}$ from an arbitrary speaker. Finally, we can obtain the facial motion with style from speaker $i$: $\hat{\mathbf{f}}=\mathcal{G}_f(\mathcal{Q}_f(\mathbf{p}_{f,i} + \Delta \hat{\mathbf{z}}_f))$ and dynamic wrinkles with style from speaker $j$: $\hat{\mathbf{w}}=\mathcal{G}_w(\mathcal{Q}_w(\mathbf{p}_{w,j} + \Delta \hat{\mathbf{z}}_w)$. 
\section{Experiments}
\subsection{Implementation Details}
\noindent \textbf{Dataset.} TexTalk4D consists of 100 audio-mesh-texture paired sequences from different subjects, each lasting around 1 minute. We split our dataset into a training set TexTalk4D-Train which contains 80-minute-long sequences from 85 subjects and a validation set TexTalk4D-Val that contains 5-minute-long sequences from 5 subjects. Following the testset split method of previous works~\cite{faceformer, facediffuser}, we constructed a \textbf{TexTalk4D-Test-A}, consisting of 5-minute-long \textbf{unseen} sequences from 10 \textbf{seen} subjects, and \textbf{TexTalk4D-Test-B}, consisting of 10-minute-long \textbf{unseen}  sequences from 10 \textbf{unseen} subjects. We use TexTalk4D-Test-A for computing objective metrics and measuring animation accuracy, and TexTalk4D-Test-B is used for perception user study and style controllability analysis. \\
\noindent \textbf{Implementation Details.} All implementations are based on PyTorch~\cite{pytorch} and NVIDIA 3090 GPUs. We use the Adam~\cite{adam} optimizer for all training. Although our texture is at 8K resolution, we downsample them to 512 for learning. The 512 motion map is sufficient to represent the motion of 8280 vertices, while for textures, it inevitably leads to loss of details. Therefore, we finetune a super-resolution network~\cite{wang2021real} to upsample the 512 wrinkle map to 8K and then combine it with the neutral 8K texture. For animation primitives learning, we train the quantized autoencoders for 500K iterations with an initial learning rate of 8e-5 and a batch size of 8.  The codebook size is $1024$ and the latent code size is $16\times16\times16$. We set $\eta_{per}=1.0$, $\eta_{adv}=0.2$ and $\eta_{code}=1.0$.  Our LDM consists of an eight-layer transformer decoder with eight attention heads. For LDM training, we set the window size $T_w=90$,$T_p=10$, diffusion steps $N=500$, and the feature dimension is 1024. The training takes 100K iterations, with a learning rate of 1e-4 and a batch size of 16. In the experiments, we train all baselines on our TexTalk4D except for DiffPoseTalk. For the training of our method and baselines, we use a HuBERT~\cite{hubert, TencentGameMate} pre-trained on Chinese data\footnote{https://huggingface.co/TencentGameMate/chinese-hubert-base} as the audio encoder for fair comparisons.
\subsection{Mesh Quality Evalution}
\noindent \textbf{Baseline.} We compare the quality of generated facial motions of TexTalker with four state-of-the-art methods: FaceFormer~\cite{faceformer}, CodeTalker~\cite{codetalker}, FaceDiffuser~\cite{facediffuser} and DiffPoseTalk~\cite{diffposetalk} on TexTalk4D-Test-A. Notably, since DiffposeTalk is only compatible with Flame~\cite{flame} parameters, we only conduct visual comparisons with it by running the public-released pre-trained model. The Flame's shape and expression coefficients are obtained by the off-the-shelf Flame tracking method~\cite{VHAP, gaussianavatars} to extract the speaking style with its style encoder.\\
\noindent \textbf{Quantitative Comparisons.} Following the previous works~\cite{codetalker, voca, meshtalk}, we measure the motion quality from three metrics: (1) Lip Vertex Error (LVE) that measures the lip-sync quality by computing the mean maximal L2 error of all lip vertices across the whole sequence, (2) Mean Full-face Vertex Error (MVE) that is similar to LVE but measures the full-face deviation compared with ground truth, and (3) Upper-face
Dynamics Deviation (FDD) that evaluates the style-relevant upper-face motions by comparing the motion standard deviation of each upper-face vertex. 

\begin{table}[t]
\resizebox{\columnwidth}{!}{%
\begin{tabular}{lccc}
\toprule
\multirow{2}{*}{Method}           & LVE$\downarrow$ & MVE$\downarrow$ & FDD$\downarrow$ \\ & $10^{-2}$mm & $10^{-2}$mm & $10^{-3}$mm\\ \toprule
FaceFormer~\cite{faceformer}       & 1.80                   & 2.94                   & 1.68                   \\
CodeTalker~\cite{codetalker}       & 1.83                   & 2.80                   & \underline{1.38}                   \\
FaceDiffuser~\cite{facediffuser}     & \underline{1.53}                   & \underline{2.38}                   & 1.64                   \\
TexTalker (Ours) &  \textbf{1.49}                      & \textbf{2.34}                       & \textbf{1.20}                       \\ \bottomrule
\end{tabular}%
}
\caption{\textbf{Facial motion comparison.} We compare our method with other SOTA methods on TexTalker4D-Test-A. Best and second-best results are marked as \textbf{bold}
 and \underline{underlined}.}
 \vspace{-1em}
\label{tab:motion_cmp}
\end{table}
Quantitative comparisons are presented in Tab.~\ref{tab:motion_cmp}. Our method outperforms FaceFormer and CodeTalker significantly in all metrics, and though we only have a slight advantage over FaceDiffuser in LVE and MVE, we surpass its FDD considerably. The results indicate that TexTalker can effectively capture the subject's facial movement style while achieving more accurate lip motions.\\
\noindent \textbf{Qualitative Comparisons.}\begin{figure}[t]
  \centering
  \includegraphics[width=\linewidth]{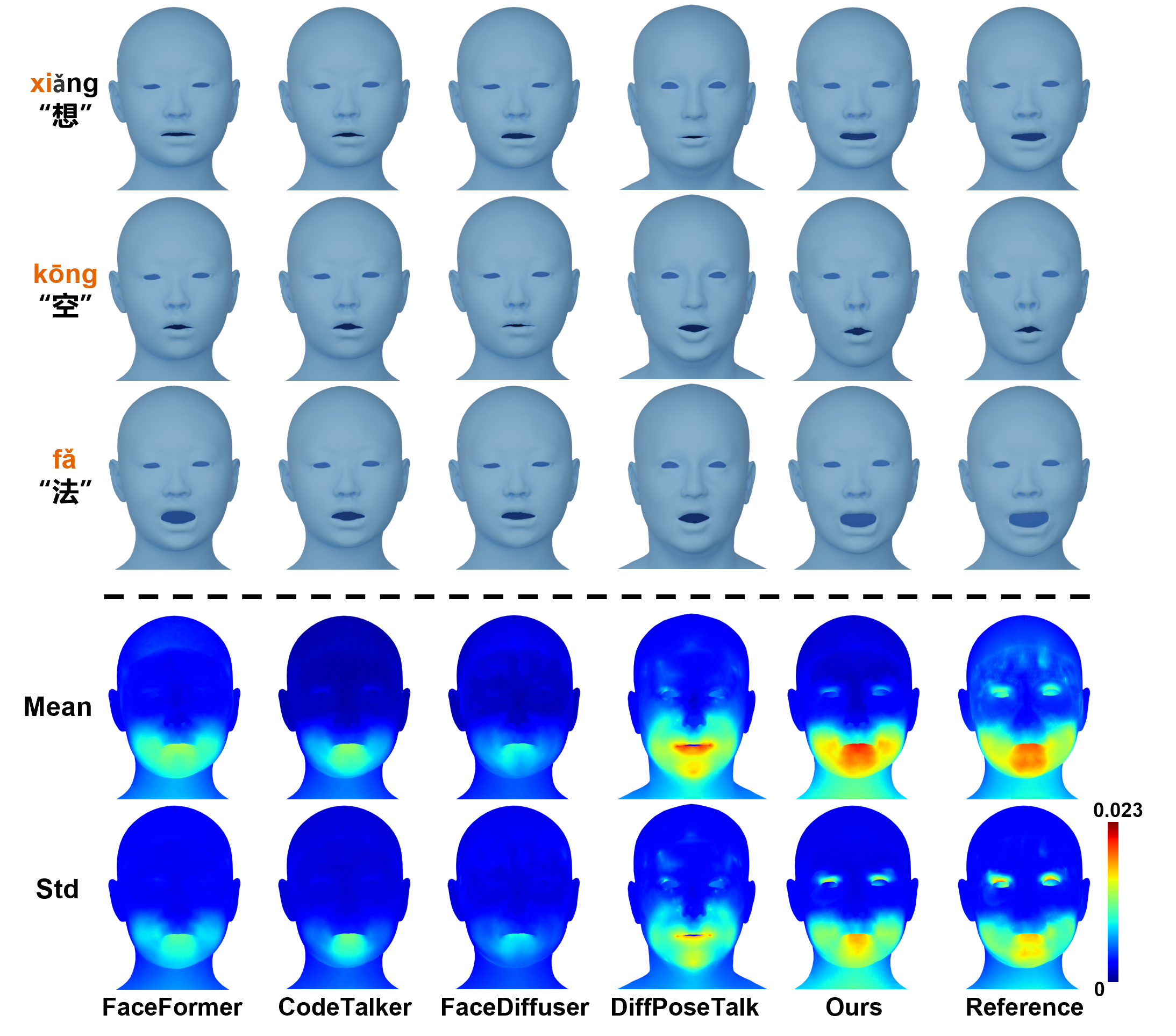}
  \caption{\textbf{Visual comparison of generated motion.} The upper partition shows samples conditioned by different phonemes and the syllables are highlighted in orange. The lower partition depicts the temporal motion statistics of the whole sequence, where the brighter the color, the more motion is observed.}
  \label{fig:motion_cmp}
\end{figure} As the visual comparisons between the facial motion shown in Fig.~\ref{fig:motion_cmp}, 
our method produces more phoneme-synced lip motions and is more consistent with the reference. The lower partition of Fig.~\ref{fig:motion_cmp} shows the facial motion dynamics by computing the temporal statistics between adjacent frames across the sequence~\cite{codetalker}. FaceFormer, CodeTalker, and FaceDiffuser are prone to generate small facial movements, and the motions are limited around the lips. Our method produces more distinct results with realistic upper-face animations, including challenging phoneme-irrelevant eye-blinking.

\subsection{Texture Quality Evaluation}
\noindent \textbf{Baseline.} 
We compare the quality of generated textures with~\cite{li2020dynamic}, where the texture is generated by an image translation method. Following their practice, we train a specially designed Pix2PixHD~\cite{p2phd} to generate dynamic textures from the neutral texture, neutral geometry map, and facial motion maps. Specifically, we train a variant of our method that only learns to generate facial motions to obtain the motion maps. In addition, to demonstrate the impact of dynamic textures on improving rendering realism, we also compared our results with static textures.\\
\noindent \textbf{Quantitative comparisons.}
\begin{table}[t]
\resizebox{\columnwidth}{!}{%
\begin{tabular}{lccclcc}
\toprule
\multirow{2}{*}{Method}      & \multicolumn{3}{c}{TexTalker4D-Test-A} &  & \multicolumn{2}{c}{TexTalker4D-Test-B} \\ \cline{2-4} \cline{6-7} 
                             & PSNR$\uparrow$        & SSIM$\uparrow$        & LPIPS$\downarrow$        &  & Realism$\uparrow$          & Consistency$\uparrow$         \\ \toprule
Static Texture & 39.79            & 0.967            & 0.0146           &  & 3.10                 & 2.65                    \\
Li~et~al.~\cite{li2020dynamic} & 42.34            & 0.981            & 0.0187           &  & 3.91                 & 3.78                    \\
Ours                         & \textbf{44.13}            & \textbf{0.985}            & \textbf{0.0101}           &  & \textbf{4.13}                 & \textbf{3.97}                    \\ \hline
\end{tabular}%
}
\caption{\textbf{Facial texture comparison.} We conduct the objective metric comparison on TexTalker4D-Test-A, and a user study on TexTalker4D-Test-B. The best results are marked as \textbf{bold}.}
\vspace{-1em}
\label{tab:tex_cmp}
\end{table} As is shown in Tab.~\ref{tab:tex_cmp}, we quantitatively compare different methods on 
TexTalk4D-Test-A using PSNR, SSIM, and LPIPS. Our method significantly outperforms Li~et~al. in all metrics. It indicates that compared with translating motion to texture, joint learning the correlation between motions and texture variations can better capture the subtle facial changes. \\
\noindent \textbf{Qualitative comparisons.}\begin{figure}[t]
  \centering
  \includegraphics[width=\linewidth]{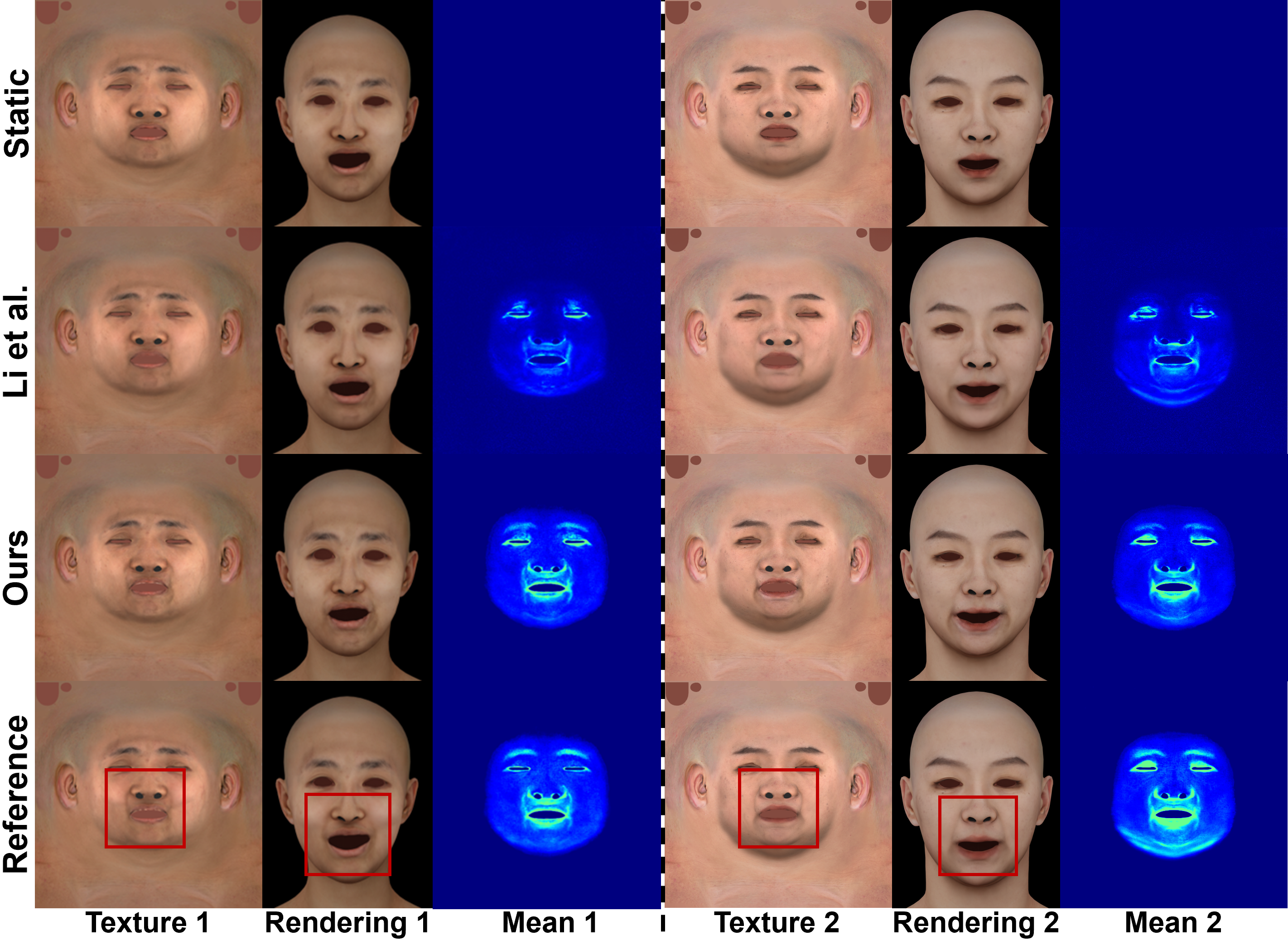}
  \caption{\textbf{Visual comparison of generated textures.} Examples of facial textures generated by different methods. Columns 3 and 6 depict the temporal texture statistics of the whole sequence, where the brighter the color, the more texture change is observed. Please zoom-in for detailed observation.}
  \vspace{-1em}
  \label{fig:tex_cmp}
\end{figure} Fig.~\ref{fig:tex_cmp} shows a visual comparison between the generated dynamic textures and the static textures. We show two example frames and the texture variation dynamics from two unseen talking sequences in TexTalk4D-Test-B. Compared with Li~et~al.~\cite{li2020dynamic}, our method generates more natural wrinkles that are consistent with underlying facial motions. Our method can even capture the challenging subtle wrinkle changes under the nose, achieving realistic rendering results. Notably, it can be observed that only using static textures for rendering will significantly reduce the realism of rendering, because the facial wrinkles reflect the strength of muscle compression related to the syllables, and the absence of texture variation leads to ambiguity in facial motions. 

The 3rd and 6th columns in Fig.~\ref{fig:tex_cmp} depict the temporal dynamic statistics of the whole sequence by computing the mean variation between adjacent frames. The image translation method tends to produce mild results and fails to reflect the strength of wrinkles faithfully. In contrast, our method produces a broader range of wrinkle dynamics and is more consistent with the reference.
\subsection{Geometry-Texture Consistency User Study}Due to the difficulty in quantifying the alignment of texture and geometry, we design a user study to compare the realism of generated textured animations and the geometry-texture consistency. Specifically, the study consists of 30 questions, each presenting the participants with a 20-second-long clip of the ground truth animation alongside those generated by different methods. Participants are asked to rate on a scale of 1 to 5 according to the realism of the animation and the consistency between texture and geometry, referring to the ground truth. As is demonstrated in Tab.~\ref{tab:tex_cmp}, our method produces the most realistic textured animation and excels at maintaining geometry-texture consistency, achieving the best perceptual quality. 
\subsection{Effect of Disentangled Style Control}\begin{figure}[t]
  \centering
  \includegraphics[width=\linewidth]{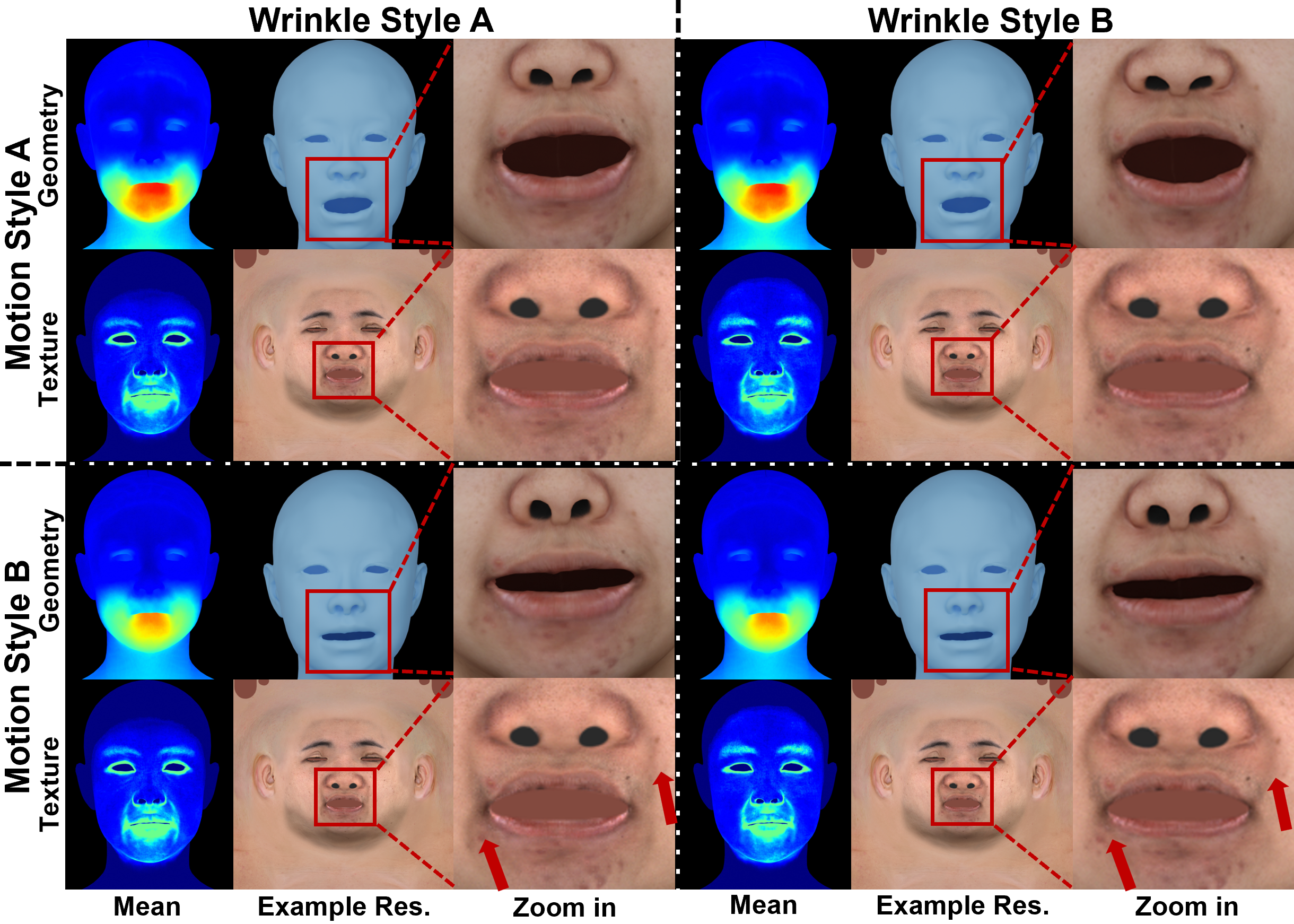}
  \caption{\textbf{Demonstration of disentangled style control.} We present dynamic statistics of the generated geometries and textures under different style conditions, with an example showcasing the texture and rendering. The red arrows highlight differences in wrinkle patterns.}
  \label{fig:stylize}
\end{figure} To explore the effect of the proposed style control method, we choose two motion-style pivots and wrinkle-style pivots, and combine them arbitrarily to control the generation. Fig.~\ref{fig:stylize} shows the visualization results under different conditions, demonstrating that the style pivots effectively capture distinct style features. By learning the offsets from style pivots, our generation network learns the phoneme-relevant but style-free facial change rules and achieves independent control over speaking and wrinkling styles, which is useful for achieving highly customized facial animation. Please refer to the Sup.~Mat. for dynamic results.

\subsection{Ablation Study}\begin{table}[t]
\resizebox{\columnwidth}{!}{%
\begin{tabular}{lccccccc}
\toprule
\multirow{3}{*}{Method}        & \multicolumn{3}{c}{Motion}                                                                                                  &                      & \multicolumn{3}{c}{Texture}                                        \\ \cline{2-4} \cline{6-8} 
                               & LVE$\downarrow$                                     & MVE$\downarrow$                                     & FDD$\downarrow$                                     &                      & PSNR$\uparrow$                 & SSIM$\uparrow$                 & LPIPS$\downarrow$                  \\
                               & \multicolumn{1}{l}{$10^{-2}$mm} & \multicolumn{1}{l}{$10^{-2}$mm} & \multicolumn{1}{l}{$10^{-3}$mm} & \multicolumn{1}{l}{} & dB & - & - \\ \toprule
Ours w/o W                     & 1.73                                        &  2.76                                       & \textbf{1.06}                                        &                      & -                    & -                    & -                    \\
Ours w/o M                     & -                                       & -                                       & -                                       &                      & \underline{43.87}                     & \textbf{0.985}                     & \underline{0.0102}                     \\
\multicolumn{1}{l}{Ours Joint} & 1.71                   & 2.68                  & 1.21                   & \multicolumn{1}{l}{} & 43.45 & 0.981 & 0.0109 \\ \multicolumn{1}{l}{Ours w/o Pivot} & \underline{1.70}                   & \underline{2.60}                  & \underline{1.18}                   & \multicolumn{1}{l}{} & 43.61 & \underline{0.984} & 0.0103 \\
Ours                           & \textbf{1.49}                                        & \textbf{2.34}                                        & 1.20                                        &                     &\textbf{44.13}                     & \textbf{0.985}                     & \textbf{0.0101}                     \\ \bottomrule
\end{tabular}%
}
\caption{\textbf{Ablation study.} (1) motion-texture joint learning v.s. separated learning, (2) motion-texture separated codebooks v.s. joint codebook, and (3) pivot-based style control v.s. one-hot style control. We compare the metrics on TexTalker4D-Test-A.}
\vspace{-1em}
\label{tab:cmp_abl_colearning}
\end{table}

In this section, we assess the impact of the crucial settings involved in our method. We conduct ablation studies in terms of \textbf{(1)} the benefit of motion-wrinkle joint learning, \textbf{(2)} the influence of facial animation primitive learning, and \textbf{(3)} the effect of pivot-based style injection. The quantitative comparison results are demonstrated in Tab.~\ref{tab:cmp_abl_colearning}.\\
\noindent \textbf{Analysis of motion-wrinkle Co-learning.} To study the effect of motion-wrinkle joint learning, we train the ``Ours~w/o~W" and ``Ours~w/o~M" that only generate motions and wrinkles, respectively. As shown in Tab.~\ref{tab:cmp_abl_colearning}, separate generation can achieve better FDD and steady SSIM, but there is a significant degradation in other metrics. It validates that separate learning may reduce multi-modal ambiguity, but joint learning can greatly promote accuracy.\\
\noindent \textbf{Analysis of Animation Primitive Learning.} To investigate whether using a single latent space to jointly store motion and wrinkle animation primitives would get better results, we train a joint quantized autoencoder with latent feature length doubled and then train our model based on the learned space (denoted as Ours-Joint). Except for FDD, all other metrics get worse, indicating that a single latent space struggles to bridge the gap between different models and using independent spaces to represent motion and wrinkle leads to better expressiveness.\\
\noindent \textbf{Analysis of Pivot-Based Style Injection.} To examine the effect of the proposed pivot-based style injection, we remove the style pivots (Ours w/o Pivot) and train the network as Eq.~\ref{equ: diffusion}, using the one-hot embedding as the style condition. The results reveal that compared to learning the offset from pivot, directly learning latent codes conditioned by one-hot makes it difficult to achieve the accurate generation of subtle mouth motions and texture variations.

\section{Conclusion}
In this paper, we propose a new task: audio-driven 3D talking head generation with corresponding dynamic textures. To address the current gap in academic datasets for this task, we present TexTalk4D, a high-precision 4D audio-mesh-texture-aligned dataset. It contains 100-minute-long LightStage reconstruction sequences from 100 subjects, with 8K textures featuring pore-level details, and spans diverse styles and audio content. Based on the dataset, we propose TexTalker to jointly generate dynamic geometry and texture under audio guidance. By compressing facial animation into low-dimensional spaces and performing the diffusion process to learn animation primitives, our method can achieve realistic animation generation with high geometry-texture consistency, outperforming existing SOTA works. Additionally, benefiting from the proposed pivot-based style injection method, our method excels in capturing complex speaking and wrinkling styles, advancing towards highly personalized textured facial animation. 

\noindent\textbf{Limitation.} Although our method can generate vivid facial animation and generalize well to other languages, races, and ages (please refer to the Sup.~Mat.), it still has some limitations. Limited by capture conditions, TexTalk4D lacks diversity in age and race. We believe that collecting more varied data will further improve generalization ability. 

{
    \small
    \bibliographystyle{ieeenat_fullname}
    \bibliography{main}
}

\clearpage

\appendix
\renewcommand{\theequation}{\Alph{equation}}
\renewcommand{\thefigure}{\Alph{figure}}
\renewcommand{\thetable}{\Alph{table}}
\setcounter{equation}{0}
\setcounter{figure}{0}
\setcounter{table}{0}

\renewcommand{\thetable}{\Alph{table}}
\renewcommand{\thefigure}{\Alph{figure}}
\renewcommand\thesection{\Alph{section}}

In the supplementary material, we provide further information about our work, including more details about our implementation (Sec.~\ref{supp:sec:imp}), the setup of our data capture system (Sec.~\ref{supp:sec:capture}), the specific design of our user study (Sec.~\ref{supp:sec:user}) with additional experiments (Sec.~\ref{supp:sec:exp_supp}), and discussions on ethical impacts (Sec.~\ref{supp:sec:ethics}). We also present a short description of our supplementary video (Sec.~\ref{supp:sec:video}). The code and dataset will be publicly released after publication.
\section{Method Details}\label{supp:sec:imp}
\subsection{Implementation Details}
The training of TexTalker consists of two parts: (1) training two codebooks that store the animation primitive of facial motion and wrinkle, respectively, and 2) training the motion-wrinkle latent diffusion model (LDM) based on the learned latent spaces. We provide further details about the implementation of the two parts. \\
\noindent \textbf{Animation Primitive Learning.} We utilize Face3D~\cite{Face3D} to map vertex offsets to UV space, generating motion maps. These maps are then normalized for better learning. For the winkle map, we use sigmoid to transform the value range to 0-1 after computing the pixel-wise ratio with the neutral texture. Taking the motion autoencoder for example, following VQGAN~\cite{vqgan}, our encoder $\mathcal{E}_f$ and generator $\mathcal{G}_f$ respectively consists of 5 resizing layers and 12 residual blocks. A single attention layer is applied on the lowest resolution. The codebook size is set to 1024 and the dimension of each item is set to 16, balancing the expressiveness and computational costs. We use the patch-based discriminator $\mathcal{D}_f$ as in ~\cite{isola2017image}. For better training stability, the discriminator is introduced after 40K iterations. All setting remains the same for the wrinkle autoencoder $\mathcal{E}_w$, $\mathcal{G}_w$ and $\mathcal{D}_w$.\\
\noindent \textbf{Motion-Wrinkle LDM.} The latent diffusion model consists of an eight-layer transformer decoder with eight attention heads. The diffusion timestep $\mathbf{n}$, clean sample $\mathbf{X}^{\prime 0}_{-T_p:0}$ from the previous window, and current noisy sample $\mathbf{X}^{\prime n}_{0:T_w}$ are first embedded into a dimension of 1024 and then concatenated together before denoising. A positional embedding layer is then applied to the features. We align the length of audio features with visual frames through linear interpolation. Following ~\cite{diffposetalk}, the learnable start features are used to generate the initial window. In addition, considering that the window length is not fixed during inference, we randomly truncate the input samples during training to improve the robustness.

\subsection{Inference Speed} We conduct inference with an NVidia 3090 GPU. Our inference stage involves two steps: (1) using LDM to generate motion and wrinkle latent codes from audio, and (2) generating motion maps and wrinkle maps separately using generator $\mathcal{G}_f$ and $\mathcal{G}_w$. Our LDM can generate motion and wrinkle latent features at 24 FPS and the generators can reconstruct maps at 38 FPS.
\section{Capture System Setup}\label{supp:sec:capture}\begin{figure}[t]
  \centering
  \includegraphics[width=\linewidth]{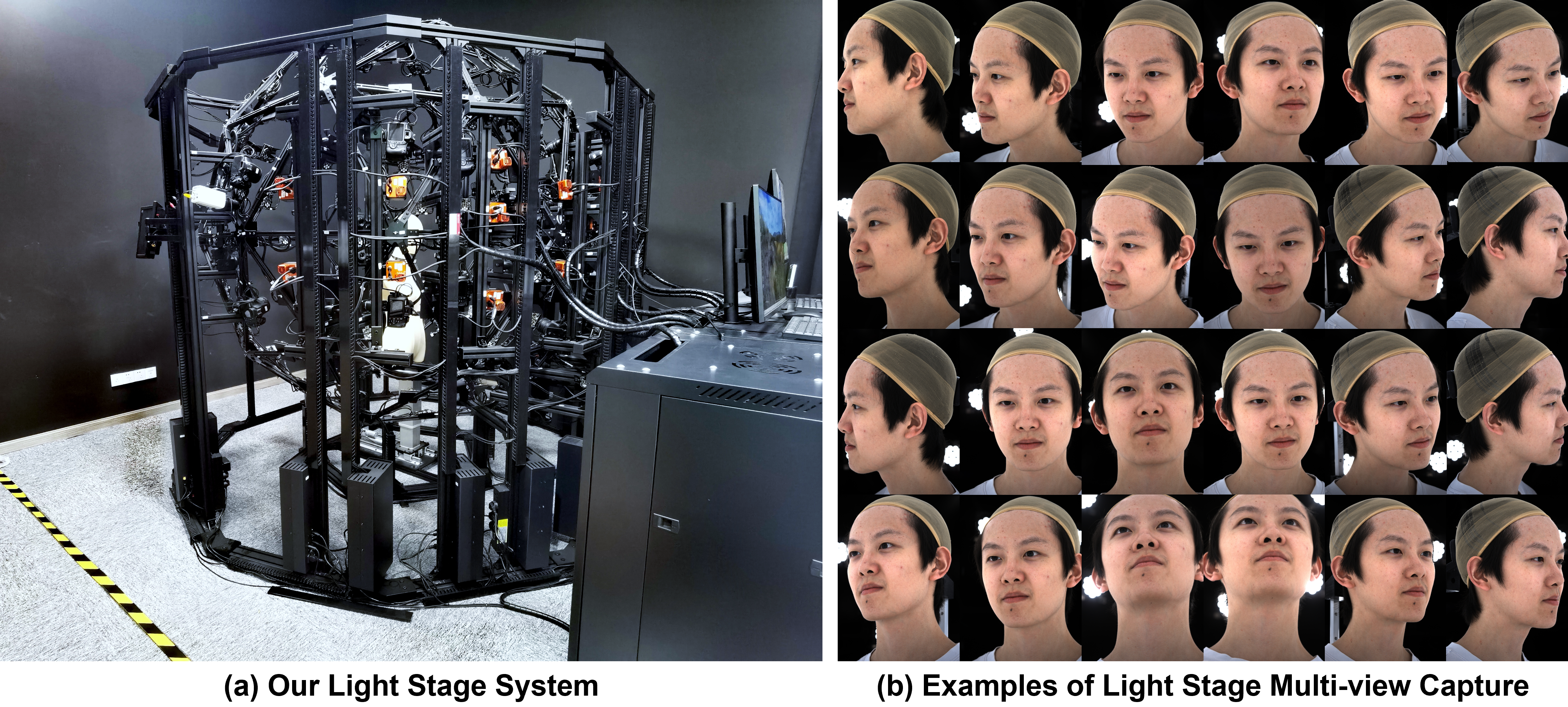}
  \caption{The diagram of our capture system.}
  \label{fig:lightstage}
\end{figure}
As is shown in Fig.~\ref{fig:lightstage}, our LightStage system for 4D data capture consists of 24 time-consistent dynamic cameras capable of capturing multi-view videos at 60FPS with a resolution of 4096 × 3000 pixels. Each camera is hardware-controlled with a time error of less than 1 microsecond.
The 24 cameras are precisely calibrated to obtain accurate intrinsic and extrinsic parameters. 
To capture real facial wrinkles while avoiding environmental influences, we employ multiple surrounding light sources for uniform lightning.
We use a dual channel 48000Hz microphone synchronized with the cameras to record talking audio.

We capture a 1-minute-long video for each of the 100 subjects. All subjects are required to prepare the text content for talking in advance. The content is reviewed by on-site staff which should have a variety of syllables and not contain any sensitive information, personal privacy, or insulting and discriminatory content. During filming, subjects are constrained to chairs while talking, maintaining relatively static body movements, which aligns with the facial capture requirements in industrial production processes. The subjects spontaneously recite the lengthy text to simulate natural speech conditions.
\section{User Study Details}\label{supp:sec:user}\begin{figure}[t]
  \centering
  \includegraphics[width=\linewidth]{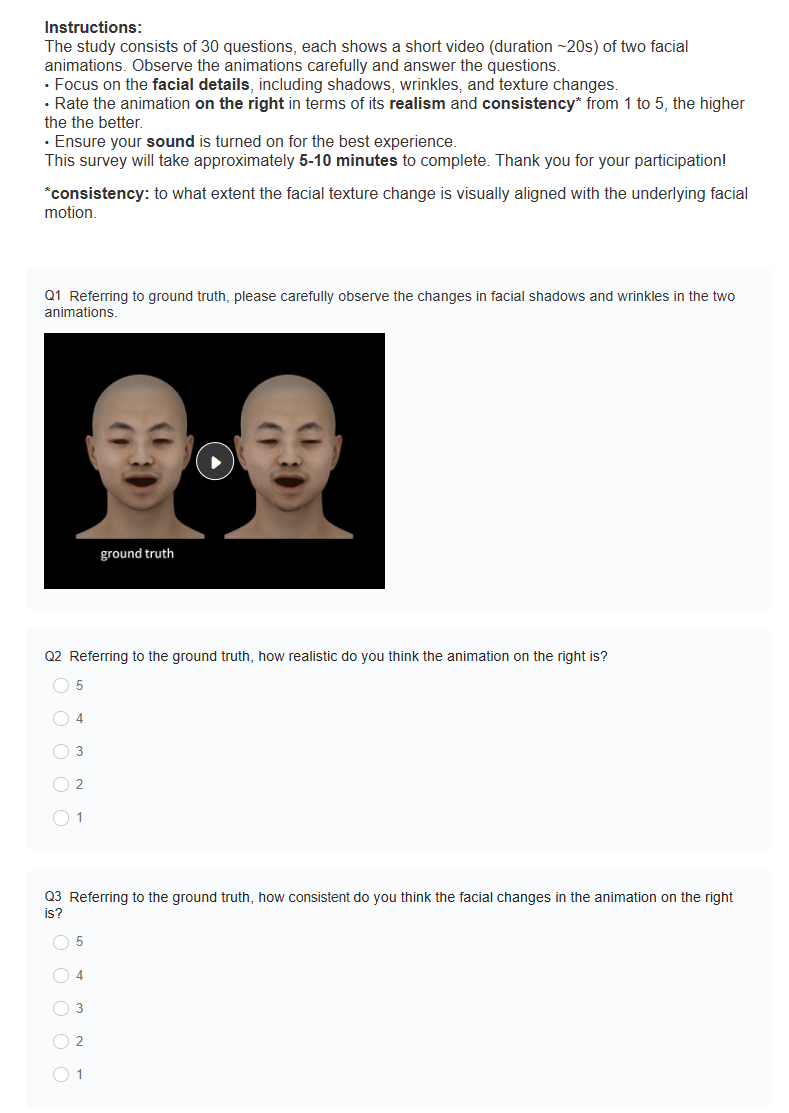}
  \caption{Demonstration of the user study interface design. }
  \label{fig:userstudy}
\end{figure}
Fig.~\ref{fig:userstudy} shows the designed user study interface. The study consists of 30 questions, each presenting the participants with a random 20-second-long clip of the ground truth animation alongside the corresponding fragments generated by an anonymous method. The order of appearance of different methods is disrupted. For each video, participants are instructed to rate the anonymous animation on a scale of 1-5 based on the ground truth, with higher ratings indicating better performance. Each question consists of two sub-items: ``Referring to the ground truth, how realistic do you think the animation on the right is?" and ``Referring to ground truth, how consistent do you think the facial changes in the animation on the right are?" To ensure the participants are fully engaged in the user study, only when they watch the entire video and rate all the items can they proceed to the next question, otherwise a warning message would pop up. Only when all questions are answered will they be included in the final results.
\section{Additional Experiments}\label{supp:sec:exp_supp}
\subsection{Additional Comparison with EmoTalk~\cite{peng2023emotalk}}
\begin{figure}[t]
  \centering
  \includegraphics[width=\linewidth]{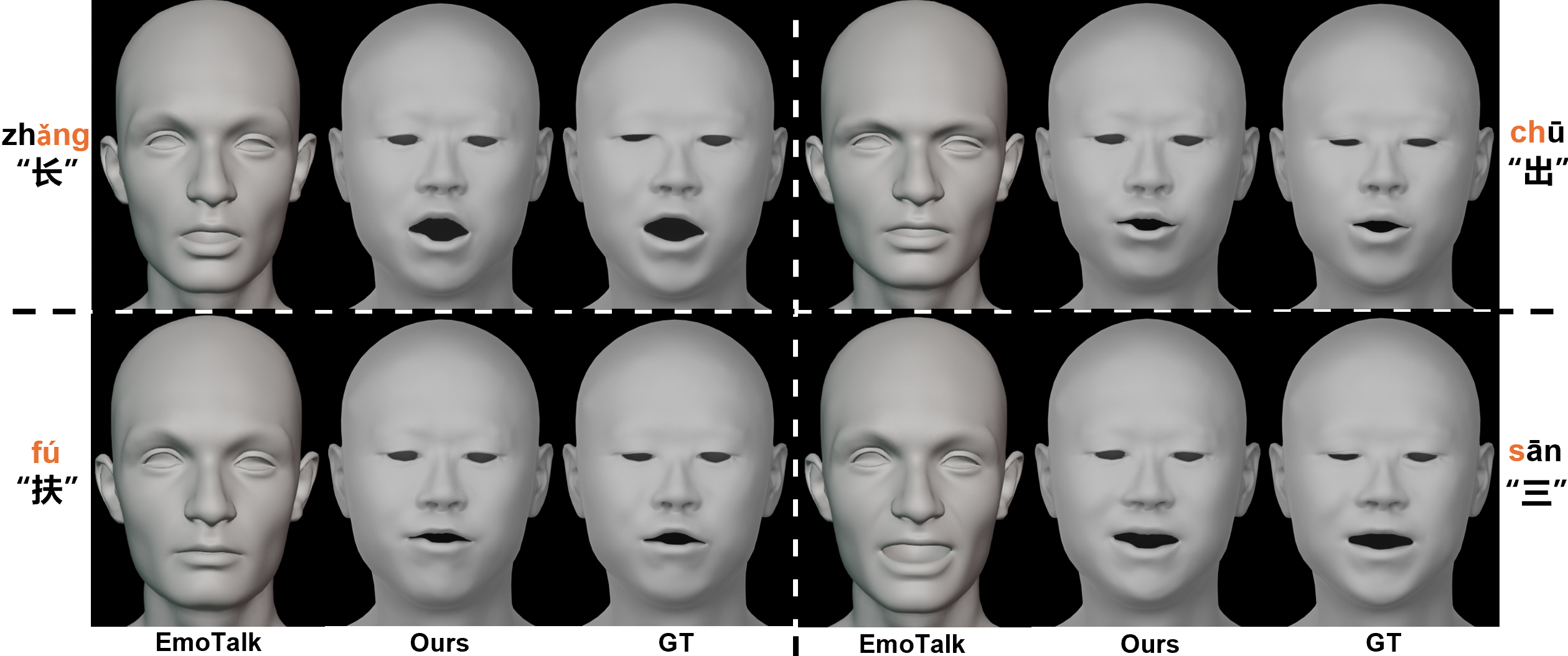}
  \caption{Comparison with EmoTalk on unseen ID and speech.}
  \vspace{-1em}
  \label{fig:emotalk_cmp}
\end{figure}
We further qualitatively compare our method with the blendshape-based method EmoTalk, using the style from the unseen identity.  As is shown in Fig.~\ref{fig:emotalk_cmp}, our method performs better than EmoTalk. Besides, EmoTalk requires artists to produce blendshapes manually and does not involve dynamic textures.
\subsection{Additional Ablation Study of Audio Encoders}
\begin{table}[t]
\resizebox{\columnwidth}{!}{%
\begin{tabular}{lccccccc}
\toprule
\multirow{3}{*}{Audio Encoder}        & \multicolumn{3}{c}{Motion}                                                                                                  &                      & \multicolumn{3}{c}{Texture}                                        \\ \cline{2-4} \cline{6-8} 
                               & LVE$\downarrow$                                     & MVE$\downarrow$                                     & FDD$\downarrow$                                     &                      & PSNR$\uparrow$                 & SSIM$\uparrow$                 & LPIPS$\downarrow$                  \\
                               & \multicolumn{1}{l}{$10^{-2}$mm} & \multicolumn{1}{l}{$10^{-2}$mm} & \multicolumn{1}{l}{$10^{-3}$mm} & \multicolumn{1}{l}{} & dB & - & - \\ \toprule

MFCC                     & 2.13                                       & 2.79                                       & 1.57                                       &                      & 43.92                     & \underline{0.984}                      & 0.0106                   \\
\multicolumn{1}{l}{Wav2Vec2-CN} & 2.10                   & 2.85                  & 1.23                  & \multicolumn{1}{l}{} & 43.64 & \underline{0.984} & 0.0105\\ \multicolumn{1}{l}{HuBERT} & \underline{1.57}                   & \underline{2.43}                  & \underline{1.21}                   & \multicolumn{1}{l}{} & \underline{44.02} & \textbf{0.985} & \underline{0.0102} \\
HuBERT-CN (Ours)                           & \textbf{1.49}                                        & \textbf{2.34}                                        & \textbf{1.20}                                        &                     &\textbf{44.13}                     & \textbf{0.985}                     & \textbf{0.0101}                     \\ \bottomrule
\end{tabular}%
}
  \vspace{-1em}
\caption{Ablation study on audio encoders. ``CN'' means model pretrained on Chinese data.}
\label{tab:abl_audio}
\end{table}

We present an additional ablation study to show the influence of different audio encoders. Specifically, we quantitatively compare the metrics of models using MFCC, Wav2Vec2-CN\footnote{https://huggingface.co/TencentGameMate/chinese-wav2vec2-base}, HuBERT and HuBERT-CN\footnote{https://huggingface.co/TencentGameMate/chinese-hubert-base}, where ``CN'' means encoder trained on Chinese data. Compared to other audio encoders, HuBERT can significantly improve generation quality. Meanwhile, encoders trained specifically on Chinese data can achieve slightly better results compared to the conventional version.
\subsection{Generalization Discussion}
\begin{figure}[t]
  \centering
  \includegraphics[width=\linewidth]{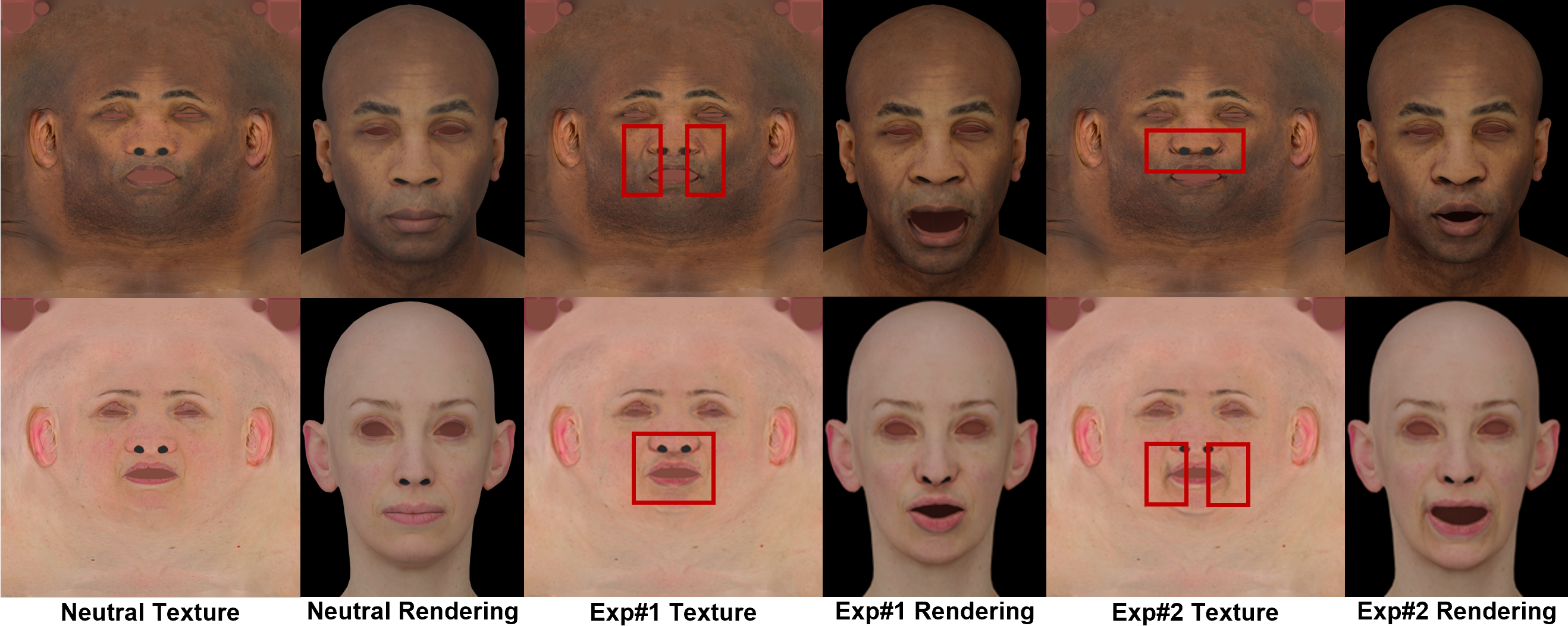}
  \caption{Results on unseen races with styles from unseen IDs.}
  \label{fig:generalization}
\end{figure}
As discussed in the limitation, our TexTalk4D dataset is mainly composed of Asian youths and only includes Mandarin speech. To study the generalization ability of the trained model, we directly test on unseen races, as is shown in Fig.~\ref{fig:generalization}. Although our dataset is limited in diversity, the model generalizes well to other languages, races, and ages. Please refer to the Supp. Video for dynamic results.

\section{Ethics Discussion}\label{supp:sec:ethics}
In our dataset, all subjects have signed agreements authorizing us to use the collected data for research. We are committed to privacy protection to prevent the misuse of the collected data for criminal purposes. Specifically, we will only disclose the reconstructed assets and not the original captures. The dataset will only be available to researchers with professional titles who apply with an official email address. Further, all subjects reserve the right to revoke the authorization, and we will stop using the relevant data, but the research already conducted will not be affected.

\section{Video Dynamic Results}\label{supp:sec:video}
To better demonstrate our work, we provide additional dynamic experimental results in the supplementary video. Specifically, we showcase several dynamic samples from our TexTalk4D dataset. For a more comprehensive evaluation of our method, we also provide more qualitative comparison results with the competitors~\cite{codetalker, facediffuser, faceformer, diffposetalk, li2020dynamic} regarding facial motion and texture generation, highlighting the superiority of our approach over previous works. Additionally, the video also shows the dynamic results of our method in the disentangled control of speaking and wrinkling styles. It proves that the proposed pivot-based style injection method can effectively capture complex styles and is useful for achieving highly personalized dynamic facial animation. Finally, we present the facial animations driven by different languages. 

\end{document}